\documentclass[journal]{IEEEtran}

\usepackage{url}

\usepackage{times}
\usepackage{epsfig}
\usepackage{graphicx}
\usepackage{amsmath}
\usepackage{amssymb}
\usepackage{algorithm,algpseudocode}

\usepackage{bm}
\usepackage{caption}

\usepackage{url}
\usepackage{booktabs}
\usepackage{multirow}
\usepackage{multirow}
\usepackage{wrapfig}
\usepackage{wrapfig,lipsum,booktabs}
\usepackage{graphicx}
\usepackage{array}
\usepackage{color, colortbl}
\usepackage{array,hhline}
\usepackage[]{hyperref}

\newif\ifshowfig\showfigtrue
\definecolor{violet}{rgb}{0.56, 0.0, 1.0}
\definecolor{cerulean}{rgb}{0.0, 0.48, 0.65}

\ifCLASSINFOpdf

   \usepackage[pdftex]{graphicx}
  \graphicspath{{../pdf/}{../jpeg/}}
  \DeclareGraphicsExtensions{.pdf,.jpeg,.png}
\else
\fi
\begin{document}
%

\title{Neighbour-level Message Interaction Encoding for Improved  Representation Learning on Graphs}
\author{Haimin~Zhang, and
        ~Min~Xu,~\IEEEmembership{Member, IEEE}

 \thanks{Haimin Zhang and Min Xu (\emph{corresponding author})  are with the School of Electrical and Data Engineering, Faculty of Engineering and Information Technology, University of Technology Sydney, 15 Broadway, Ultimo, NSW 2007, Australia (Emails: Haimin.Zhang@uts.edu.au, Min.Xu@uts.edu.au).}
 }
\maketitle

\begin{abstract}
  Message passing has become the dominant framework in graph representation learning.
  The essential idea of the message-passing framework is  to update node embeddings based on the information aggregated from local neighbours.
  However, most existing aggregation methods have not encoded neighbour-level message interactions  into the aggregated message, resulting in an information lost in embedding generation.
  And this information lost could be accumulated and become more serious as more layers are added to the graph network model.
  To address this issue, we propose a neighbour-level message interaction information encoding method for improving graph representation learning.
  For messages that are aggregated at a  node, we explicitly generate an encoding between each message and the rest messages using an encoding function.
  Then we aggregate these learned encodings and take the sum of the aggregated encoding and the aggregated message to update the embedding for the  node.
  By this way,  neighbour-level message interaction information is  integrated into the generated node embeddings.
  The proposed encoding method is a generic method which can be integrated into  message-passing graph convolutional networks.
  Extensive experiments are conducted  on six popular benchmark datasets across four highly-demanded tasks. 
  The  results show that  integrating  neighbour-level message interactions achieves improved  performance of the base models, advancing the state of the art results for  representation learning over graphs.

\end{abstract}

\begin{IEEEkeywords}
graph convolutional networks, graph representation learning, neighbour-level message interaction encoding.

\end{IEEEkeywords}

\IEEEpeerreviewmaketitle

\iftrue
\section{Introduction} \label{sec:intro}
Graph-structured data are very commonly seen throughout various domains such as neural science and social science.
Social networks,  brain functional networks, communication networks and molecular structures---all of these types of data have an underlying  graph structure.
Therefore, it is of considerable significance to develop models that can learn and generalize from these graph-structured data.
Over the past years, increasing studies have been devoted to   representation learning on graphs, including generalizations of convolutional neural networks to non-Euclidean data, techniques for geometry processing, and neural message passing  approaches \cite{hamilton2020graph}.
These efforts have produced new influential  results in a variety of domains, including recommendation systems \cite{he2020lightgcn}, drug discovery \cite{ji2021graph,sun2020graph},    2D and 3D vision \cite{xu2021deepfea,ding2021spatial}, and bioinformatics \cite{zhang2021graph}.

Basically, a graph contains  a set of nodes  and a set of edges between pairs of
these nodes \cite{hamilton2020graph}.
For instance, in as social network nodes could be used  to represent users and  edges can be used to represent  a connection between pair of users.
Unlike images and natural languages, which essentially have an fundamental   grid or sequence structure, graph-structured data have an fundamental structure that is in a non-Euclidean space.
It is a complicated task to develop models that are able to learn and generalize over these graph-structured data.
Early attempts employed recursive neural networks to process  directed acyclic-structured graph data.
Later on, researchers developed models that can deal with general graphs \cite{gori2005new,scarselli2008graph}.
These early models primarily contain a recurrent process which iteratively updates node states and exchanges information between nodes until these node states reach a stable equilibrium.

Bruna  \cite{bruna2014spectral} first derived graph convolutional network model that extends the concept of convolution in signal processing to non-Euclidean graphs.
Since then, a variety of graph neural network models have been developed, especially in recent years.
While these graph neural network models can be motivated in different ways, they can be grouped into  spectral approaches and spatial  approaches.
The spectral approaches, which  generalize the notion of  signals and  convolutions to the graph domain, define  convolutions through an extension of the Fourier transform to graphs \cite{bruna2014spectral}.
Unlike the spectral approaches, the spatial approaches directly define convolutions on  spatially localized neighbours and generate  features  for nodes according to information aggregated from local neighbourhood \cite{hamilton2017inductive,velickovic2018graph}.
Both spectral and  spatial graph convolutional networks are essentially  neural networks \cite{balcilar2020analyzing} that  utilize  a  message passing paradigm.
In this paradigm,  messages are exchanged between nodes and  node representations are updated based on the messages using  neural networks.

Message passing is significant for current graph convolutional networks.
The  idea of message passing is to generate  embeddings for every nodes through aggregating information from a  local neighborhood.
The most commonly used  aggregation operation  simply takes the sum of the messages from a node's local neighborhood.
For instance, in the GCN model \cite{kipf2017semi}, features from a node's local neighbours are multiplied by a factor which equals to the reciprocal of the  node's neighbour set size and then added together to update the node's representation.
In  GatedGCN  \cite{bresson2017residual},  features from a node's local neighbours are multiplied by an edge gate vector computed by the gating mechanism  and then the gated representations are added together to update the  node's  embedding.

\iftrue
\begin{figure}[!t]

  \begin{center}

  \hspace{0pt}\includegraphics[width=.495\textwidth]{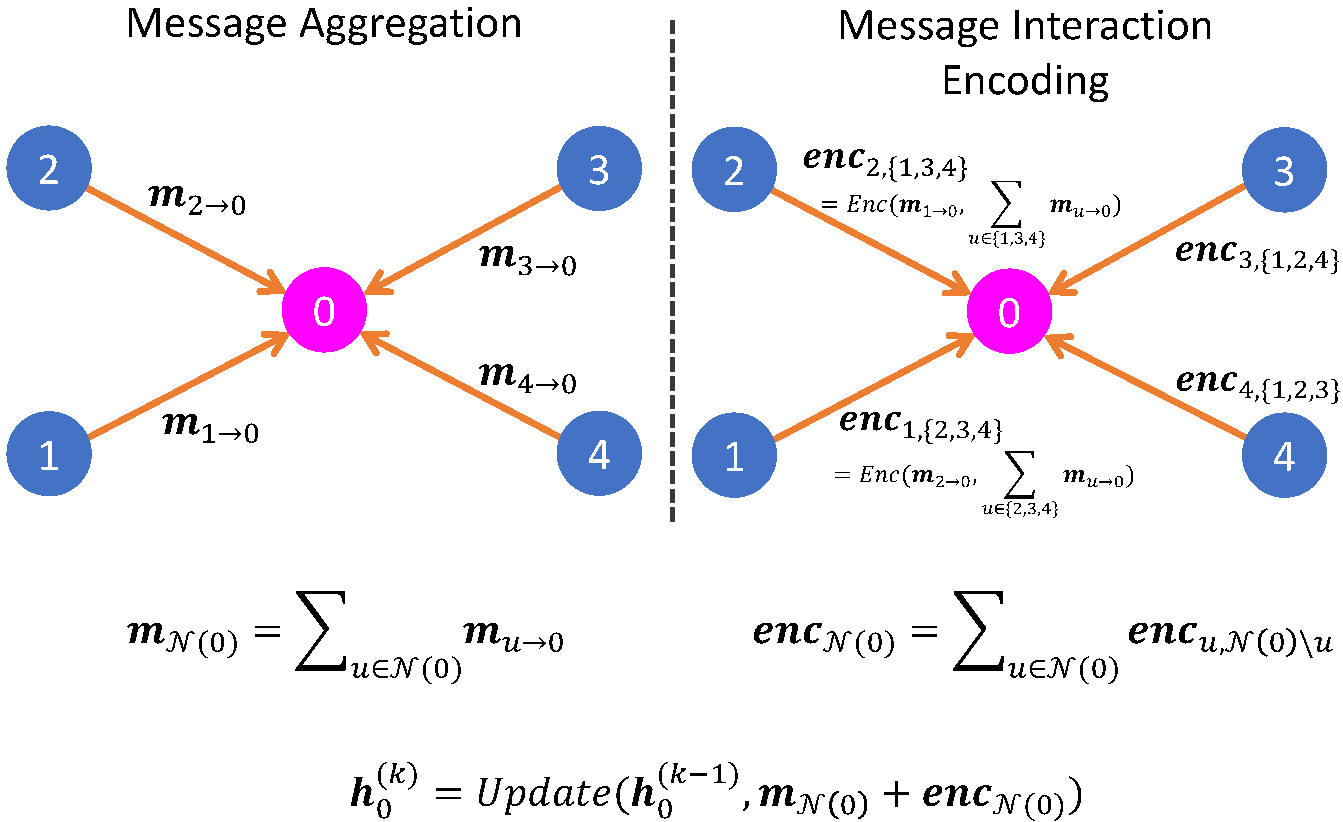} 
  \end{center}
  \caption{For each message that passes from a neighbour node to the target node, an encoding is learned between this message and the aggregated message from the rest neighbours. These learned encodings are aggregated to generate a neighbour-level message interaction encoding. The sum of the aggregated message and neighbour-level message encoding is taken to update the embedding for the target node.} 
  \label{fig:fig1}
\end{figure}
\fi

Message aggregation is a key step in the message passing framework.
However, in most existing aggregation methods the neighbour-level message interaction information is not encoded in the aggregated message.
Therefore there exists information lost at each iteration of message passing, and this information lost could be accumulated as more layers are added to the model, resulting in reduced performance in representation learning over graphs.
To deal with this issue, we propose a neighbour-level message interaction encoding method for improving graph convolutional networks.
An illustration of our method is demonstrated  in Figure \ref{fig:fig1}.
During each round of message passing,  we first follow the conventional approach to aggregate messages from a node's local neighbourhood to generate an aggregated message for the node.
Then for each message from a neighbour node, we explicitly learn an encoding between this  message and the aggregated message from the rest neighbour nodes in the local  neighbourhood.
We aggregate these learned encodings and combine the aggregated encoding with the aggregated message to generate an updated embedding for the center node of the local neighbourhood.
Because each of the learned encoding encodes the interaction information between a node and the rest nodes, the neighbour-level message interaction information is integrated into the updated embeddings.

The proposed neighbour-level message encoding method is a generic approach that can be incorporated in  message-passing graph convolutional networks.
We validate the proposed method on two base models: GCN and GatedGCN.
We carry out  experiments on six  benchmark datasets, including MNIST, CIFAR10, PATTERN, CLUSTER, TSP and ZINC \cite{dwivedi2020benchmarking}.
These experiments are carried out on four graph domain tasks, including superpixel graph classification, node classification, edge prediction and  graph regression.
The experimental results show that integrating neighbour-level message interaction information consistently improves the  performance of the base models, advancing the state of the art performance for representation learning on these datasets.



To summarize, this paper provides the following contributions.
\begin{itemize}
  \item
  This paper proposes a  neighbour-level message interaction encoding method for improving graph convolutional networks.
  For each message that passes from a neighbour node to the target node, we learn an embedding between this message and the aggregated message from the rest neighbours. We aggregate these learned encodings to generate a neighbour-level message encoding and take the sum of the aggregated message and the neighbour-level message encoding to update the target node's embedding.
  This significantly improves the representational ability of generated node embeddings.


  \item The proposed encoding method is a generic method which can be integrated in current message passing graph convolutional networks.
      It explicitly  addresses the  issue of information lost in the embedding generation process, which exists in most existing graph convolutional networks.

  \item We demonstrate that the proposed method consistently improves the performance of the base models on six benchmark datasets across four domain tasks, advancing the state of the art performance for representation learning on these benchmark datasets.

\end{itemize}

\fi   

\iftrue
\section{Related Work} \label{sec:related_work}

Generalizing neural networks to non-Euclidean data has been studied for decades.
The first-generation graph neural work models were developed by Gori et al. \cite{gori2005new} and Scarselli et al. \cite{scarselli2008graph}.
These early models  primarily use recurrent neural networks to generated node representation in an iterative manner.
This process is usually computationally expensive.
Driven by convolutional neural networks' considerable success,
the past years have seen a surge in studies on generalizing  convolutional neural networks to the graph domain.
These efforts have resulted in new theories and a variety of models for representation learning on graphs.

Most existing graph convolutional networks can be  categorized into  spectral approaches and  spatial approaches \cite{wu2020comprehensive}. 
The spectral  approaches are  motivated by the spectral graph theory.
The key feature of the spectral approaches is that they define graph convolutions through an extension of the Fourier transform to graphs.
Bruna et al. \cite{bruna2014spectral} developed the basic graph convolutional network model, in which convolutions are formulated in the Fourier domain based on the eigendecomposition of the graph Laplacian matrix.
This basic graph convolutional network model was later improved by the Chebyshev graph network (ChebNet) model \cite{defferrard2016convolutional}.
The ChebNet models constructs convolutions according to the Chebyshev expansion of the graph Laplacian, which eliminates the process for graph Laplacian decomposition and can result in spatially closed kernels.
Finally, the influential GCN model  \cite{kipf2017semi} was introduced, addressing the limitations in the previous methods.
The  GCN model is a layered architecture based on a first-order approximation of spectral  convolutions on graphs.

Unlike the spectral approaches, the spatial approaches directly defines convolutions on  graph nodes in a local neighbourhood and generate node representations by aggregating information from a local neighbourhood.
Monti et al. \cite{monti2017geometric} proposed the mixture model network (MoNet) model, which is a spatial approach that generalizes convolutional neural networks  to graphs and manifolds.
Hamilton et al. \cite{hamilton2017inductive} developed the GraphSAGE model, which generates  representations for nodes by  sampling a fixed-size set of neighbours and aggregating  features from local neighbours using an aggregation function.
Velickovic et al. \cite{velickovic2018graph} introduced to integrate the self-attention mechanism  which learns a weight for each feature from a neighbour node in the aggregation step.
This method enables the model to attend on important information in local neighbourhood feature aggregation. 
Lee et al. \cite{lee2023towards} introduced to integrate edge attention and hop attention for embedding generation and showed that this method helps to alleviate the issue of over-smoothed features and attentions.
Gao et al. \cite{gao2022patchgt} proposed to segments a graph into
patches using spectral clustering.
Then a graph neural network is used to learn patch-level representations and a Transformer model is used to generate graph-level representations.
Compared with the spectral approaches, the spatial approaches are much efficient and can be scalable  to arbitrarily structured graphs.

In recent years, there has  been a growing endeavor to  develop deeper graph network models.
Chen et al. \cite{chen2020simple} introduced the GCNII model, which extends the GCN model with initial residual connection and identity mapping.
They theoretically showed that a GCNII model with $K$ layers is able to represent a spectral filter in polynomial form up to order $K$ with arbitrary coefficients, and this property makes it possible to  build deep graph convolutional network models.
Zhao et al. \cite{zhao2020pairnorm} introduced a normalization layer called PairNorm, which maintains a consistent total pairwise feature distance across layers, in order to address the over-smoothing issue in building deeper  graph network models.
Zhang et al. \cite{zhang2022ssfg} proposed a method that stochastically scale features and gradients during training.
They showed that this method helps to improve the learning performance and generalization performance.
Xu et al. \cite{xu2018representation} developed the JKNet model, which  combines the output of each layer using  dense skip connections  to maintain the locality of node representations to preventing the over-smoothing problem.
Lee et al. \cite{lee2023towards} proposed the AERO-GNN model, which helps to address the issues of over-smoothness of embeddings and cumulative attentions in deep graph attention networks.


\fi

\iftrue
\section{Methodology}

In this section, we first describe the notations and the message passing framework.
Then we introduce the proposed neighbour-level message interaction encoding method for improving  message-passing graph convolutional networks.

\subsection{Notations}
Formally, a graph $G = (V, E)$ is  defined through a set of nodes, or called vertices, $V$ and a set of edges  $E$ between pairs of these nodes.
An edge  from node $u\in{V}$ to node $v\in{V}$ is represented as $(u,v)$.
$\mathcal{N}(u)$ denotes the set  of node $u$'s neighbouring nodes.
The adjacent matrix of the graph $G$ is denoted as $\mathbf{A} \in \mathbb{R}^{|V| \times |V|}$, in which $\mathbf{A}_{u,v}=1$ if $(u,v)\in {E}$ or $\mathbf{A}_{u,v}=0$ otherwise.
The degree matrix $\mathbf{D}$ of ${G}$ is a $|V| \times |V|$ diagonal matrix wherein $\mathbf{D}_{ii}=\sum_j \mathbf{A}_{ij}$.
The node-level feature or attribute associated with each node $u\in V$ is denoted as $\mathbf{x}_u$.

The  Laplacian matrix of $G$ is defined as  $\mathbf{L}=\mathbf{D}-\mathbf{A}$, and the symmetric normalized Laplacian is defined as $\mathbf{L}^{sym} = \mathbf{I} -\mathbf{D}^{-1/2}\mathbf{A}\mathbf{D}^{-1/2}$, where $\mathbf{I}$ is an identity matrix.
The symmetric normalized Laplacian $\mathbf{L}^{sym}$ is positive semidefinite and can be factorized as $\mathbf{L}^{sym}=\mathbf{U}\mathbf{\Lambda}\mathbf{U}^T$, where $\mathbf{\Lambda}=\text{diag}(\lambda_1,...,\lambda_{|V|})$ is a diagonal matrix of the eigenvalues and $\mathbf{U}=[\mathbf{u}_1,..., \mathbf{u}_{|V|}] \in \mathbb{R}^{|V|\times |V|}$ is a matrix of eigenvectors.

\subsection{The Message-Passing Framework}

Graph convolutional network models can be motivated in different ways.
From one perspective, the basic graph convolutional network model is derived based on the spectral graph theory,
as a generalization of Euclidean convolutions to  non-Euclidean graphs \cite{bruna2014spectral}.
The  graph convolution   is defined as the product of a signal $\mathbf{s}\in \mathbb{R}^N$ with a filter $g_\theta$ parameterized by $\bm{\theta}\in \mathbb{R}^N$ in the Fourier domain:
\begin{equation} \label{eq:edge_feature}
    \begin{aligned}
     g_{\theta}\ast \mathbf{s} =\mathbf{U}g_{\theta}^{\ast}(\mathbf{\Lambda})\mathbf{U}^T \mathbf{s},
    \end{aligned}
\end{equation}
where $\ast$ denotes the convolution operation. $g_\theta$ can be understood as a function of the eigenvalues, i.e., $g_\theta=g_{\theta}^{\ast}(\mathbf{\Lambda})$,
and $\mathbf{U}^{T}\mathbf{s}$ is the graph Fourier transform of $\mathbf{s}$.

From another perspective, graph convolutions can be defined on a spatially localized neighbourhood. The   representations for nodes are generated by aggregating information from a local  neighbourhood.
This behaviour is analogous to that of the convolutional kernels in convolutional neural networks, which aggregate features from spatially defined patches in an image.
Fundamentally,  spectral and  spatial graph convolutional networks are  neural networks \cite{balcilar2020analyzing} that  use  a message-passing form.
In this form,   messages are exchanged between graph nodes and node representations are updated based on the messages using  neural networks \cite{gilmer2017neural}.  
At each layer of message passing, a hidden representation $\mathbf{h}_{u}^{(k)}$ for each node $u\in V$ is updated according to the information aggregated from the node's local neighbourhood.
According to Hamilton \cite{hamilton2020graph}, this message passing framework can be  expressed as follows:
\begin{equation} \label{eq:message_passing}
    \begin{aligned}
     \mathbf{h}_{u}^{(k)} = Update^{(k)}(\mathbf{h}_u^{(k-1)}, Agg^{(k)}({\mathbf{h}_v^{(k-1)}}, \forall v \in \mathcal{N}(u))),
    \end{aligned}
\end{equation}
where $Update$ and $Agg$  are a differentiable function, e.g., neural networks.
The superscripts are used for differentiating the embeddings and functions at different layers of message passing.
The embeddings at $k=0$ are initialized to the node-level features, i.e., $\mathbf{h}_u^{(0)}=\mathbf{x}_u, \forall u\in V$.
During each iteration of message passing, the $Agg$ function aggregates the embeddings of nodes in $u$'s graph neighbourhood $\mathcal{N}(u)$ and generates a representation according to the aggregated neighborhood information.
The $Update$ function then generates an updated embedding for node $u$ based on the aggregated information and its previous layer embedding.
After $k$ message passing iterations, the embedding for each node  contains information from its $k$-hop neighborhood.

\begin{algorithm}[tb]
\caption{The embedding generation algorithm with the proposed neighbour-level message interaction information encoding method.}
\label{alg:1}
\hspace*{0.02in} {\bf Input:} Graph $G =(V, E)$; number of message passing iterations $K$; input node features $\{ \mathbf{x}_v, \forall u\in V \}$ \\
\hspace*{0.02in} {\bf Output:} Node embeddings $\mathbf{h}_u^{(K)}$ for all $u\in V$
\begin{algorithmic}[1]
\State {$\mathbf{h}_u^{(0)} \leftarrow \mathbf{x}_u, \forall u\in V $}
\For{$k=1,...,K$}
    \For{$u\in V$}
       \State {\color{blue} /*  aggregate messages from $u$'s local neighbours. */}
       \State $\mathbf{m}_{\mathcal{N}(u)}^{(k)} = Agg^{(k)}(\{\mathbf{h}_v^{(k-1)}, \forall v \in \mathcal{N}(u)\})$

       \State \hspace{27.5pt} $=\sum_{v \in \mathcal{N}(u)}\mathbf{m}_{v\to u}^{(k)} $ 

       \For{$v \in \mathcal{N}(u)$}
         \State {\color{blue} /* learn an encoding between the message from neighbour node $v$ and  the aggregated message from the rest neighbours. */}
         \State $\mathbf{enc}_{v,\mathcal{N}(u)-v} = fc(Concat($
         \State \hspace{27.5pt} $\mathbf{m}^{(k)}_{v\to u}, \mathbf{m}_{\mathcal{N}(u)}^{(k)} - \mathbf{m}^{(k)}_{v\to u}))$
       \EndFor
       \State {\color{blue} /* combine the aggregated message and the aggregated encoding to update the embedding for $u$. */}
       \State $\mathbf{h}_u^{(k)} = Update^{(k)}(\mathbf{h}_u^{(k-1)}, \mathbf{m}_{\mathcal{N}(u)}^{(k)} + \mathbf{enc}_{\mathcal{N}(u)})$
       \State \hspace{12.5pt} $=  Update^{(k)}(\mathbf{h}_u^{(k-1)}, \mathbf{m}_{\mathcal{N}(u)}^{(k)} + \sum_{v\in \mathcal{N}(u)} fc($
       \State \hspace{35.5pt} $Concat(\mathbf{m}^{(k)}_{v\to u}, \mathbf{m}_{\mathcal{N}(u)}^{(k)} - \mathbf{m}^{(k)}_{v\to u}))$
    \EndFor

\EndFor

\end{algorithmic}
\end{algorithm}
\subsection{Neighbour-level Message Interaction Encoding for Improving Graph Convolutional Networks}

The framework of message passing  is at the core of current graph convolutional networks and is currently the dominant framework for  representation learning over graphs.
The essential idea of the message passing framework is to update the embedding for a node based on the   message aggregated from the node's local neighbours.
Therefore this message passing update of Equation (\ref{eq:message_passing}) can also be described as follows. 


\begin{equation} \label{eq:masspassing}
    \begin{aligned}
     &\mathbf{m}_{v\to u}^{(k)} = Message^{(k)}(\mathbf{h}_v^{(k-1)}, \mathbf{h}_u^{(k-1)}) \\
     &\mathbf{m}_{\mathcal{N}(u)}^{(k)} =\sum_{v \in \mathcal{N}(u)} \mathbf{m}_{v\to u}^{(k)} \\
     &\mathbf{h}_u^{(k)} = Update^{(k)}(\mathbf{h}_u^{(k-1)}, \mathbf{m}_{\mathcal{N}(u)}^{(k)} )
    \end{aligned}
\end{equation}
For each node $v\in \mathcal{N}(u)$, a message $\mathbf{m}_{v\to u}^{(k)}$ that passes from node $v$ to node $u$ is generated with a $Message$ function.
Then an aggregated message $\mathbf{m}_{\mathcal{N}(u)}^{(k)}$ is generated by aggregating the messages from node $u$'s local neighbours.
After that, the embedding for node $u$ is updated according to the aggregated message $\mathbf{m}_{\mathcal{N}(u)}^{(k)}$ and its previous embedding $\mathbf{h}_u^{(k-1)}$.
For instance, in the GAT model, the message from a neighbour node is represented as the multiplication of the feature from this neighbour and an attention weight computed by the self-attention function, and the messages from the node's neighbours are added up together to update its embedding.

While this message passing update approach is effective, the neighbour-level message iteration information is not encoded in  generated node embeddings.
And this information lost can be accumulated as more message passing layers are added to the model, resulting in reduced performance on graph representation learning.
To address this problem, we propose a neighbour-level message interaction encoding method for improving graph convolutional network.
The idea of the our method is as follows: For each node $v \in \mathcal{N}(u)$, we learn an embedding between the message from node $v$ and the aggregated message from the rest nodes in the local neighbourhood.
Then we aggregate these learned encodings and combine the aggregated encoding and the aggregated message to update the embedding for node $u$.
Through this way, the neighbour-level message interaction information  is encoded in the generated node embeddings, and therefore the representational ability of the generated embeddings is improved.

Specifically, at each layer of message passing, we first follow Equation (\ref{eq:masspassing}) to generate a message $\mathbf{m}_{v\to u}^{(k)}$ for each $v \in \mathcal{N}(u)$ and take  the sum of the messages from $u$'s neighbours to generate an aggregated message $\mathbf{m}_{\mathcal{N}(u)}^{(k)}$.
Then for each message $\mathbf{m}_{v\to u}^{(k)}$, we learn an encoding between this message and the aggregated message from $\{\mathcal{N}(u)-v\}$, i.e., $\sum_{a \in \{\mathcal{N}(u)-v\}}\mathbf{m}^{(k)}_{a\to u}$ using a fully connected layer $fc$ as follows:

\begin{equation}
\begin{aligned}
     \mathbf{enc}_{v,\mathcal{N}(u)-v} = fc^{(k)}(Concat(\mathbf{m}^{(k)}_{v\to u}, \sum_{a \in \{\mathcal{N}(u)-v\}}\mathbf{m}^{(k)}_{a\to u})),
\end{aligned}
\end{equation}
where $Concat$ denotes the concatenation operation.
Because $\mathbf{m}_{\mathcal{N}(u)}^{(k)}=\sum_{v \in \mathcal{N}(u)} \mathbf{m}_{v\to u}^{(k)}=\mathbf{m}^{(k)}_{v\to u}+\sum_{a \in \{\mathcal{N}(u)-v\}}\mathbf{m}^{(k)}_{a\to u}$, the above equation can be reformulated as follows:
\begin{equation}
\begin{aligned}
     \mathbf{enc}_{v,\mathcal{N}(u)-v} = fc^{(k)}(Concat(\mathbf{m}^{(k)}_{v\to u}, \mathbf{m}_{\mathcal{N}(u)}^{(k)} - \mathbf{m}^{(k)}_{v\to u})).
\end{aligned}
\end{equation}
Finally we aggregate these learned encodings and combine the aggregated encoding with the aggregated message $\mathbf{m}_{\mathcal{N}(u)}^{(k)}$ to generate the updated embedding $\mathbf{h}_u^{(k)}$ as:
\begin{equation}
\begin{aligned}
     \mathbf{h}_u^{(k)} &= Update^{(k)}(\mathbf{h}_u^{(k-1)}, \mathbf{m}_{\mathcal{N}(u)}^{(k)} + \sum_{v \in \mathcal{N}(u)} \mathbf{enc}_{v,\mathcal{N}(u)-v}) \\
     &= Update^{(k)}(\mathbf{h}_u^{(k-1)}, \mathbf{m}_{\mathcal{N}(u)}^{(k)} + \mathbf{enc}_{\mathcal{N}(u)}),
\end{aligned}
\end{equation}
where $\mathbf{enc}_{\mathcal{N}(u)} = \sum_{v \in \mathcal{N}(u)} \mathbf{enc}_{v,\mathcal{N}(u)-v}$ is the aggregated encoding.
Algorithm \ref{alg:1} describes  the node embedding generation, i.e., forward propagation, algorithm with neighbour-level message interaction information encoding.
Because each of the learned encoding $\mathbf{enc}_{v,\mathcal{N}(u)-v}$ encodes the information between the message from node $v$ and the messages from the rest neigbhour nodes in the local neighbourhood, the neighbour-level message interaction information is integrated in the updated embedding $\mathbf{h}_u^{(k)}$.

The proposed neighbour-level message interaction encoding method is generic method that can be integrated into  message passing graph convolutional networks.
In this work, we validate the proposed method on two base models: the basci GCN model \cite{kipf2017semi} and the GatedGCN model \cite{bresson2017residual}.
In the following, we describe applying the proposed neighbour-level message interaction encoding method on the two base models.

\iftrue
\begin{figure*}[!t]

  \begin{center}
  \includegraphics[width=\textwidth]{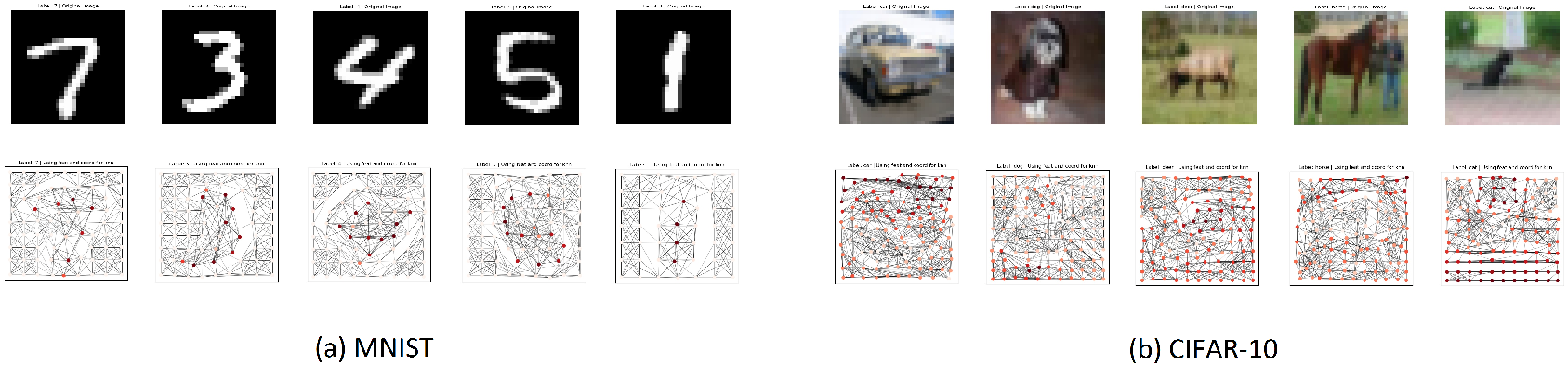} 
  \end{center}
  \caption{Examples of superpixel graphs  generated from the original images  in MNIST and CIFAR10. The node features of the superpixel graphs are the coordinates and intensity of the superpixels.}
  \label{fig:samples-mnist-cifar}
\end{figure*}
\fi

\textbf{The basic GCN}.
The basic GCN model updates the  embedding for a node by computing the average of the node's neighbouring features as follows:
\begin{equation}
\begin{aligned}
     \mathbf{m}^{(k)}_{\mathcal{N}(u)} &= \sum_{v\in \mathcal{N}(u)}\mathbf{m}_{v\to u}^{(k)}\\
       &= \sum_{v\in \mathcal{N}(u)} \frac{1}{| \mathcal{N}(u) |} W^{(k)} \mathbf{h}_v
\end{aligned}
\end{equation}
where $W^{(k)}$ is a trainable weight  matrix and $\mathbf{m}_{v\to u}^{(k)} = \frac{1}{| \mathcal{N}(u) |} W^{(k)} \mathbf{h}^{(k)}_v$ is the message that passes from node $v$ to node $u$.
With the neighbour-level message interaction encoding method, during each  message-passing iteration, we update the embedding for each $u\in V$  as follows:
\begin{equation}
\begin{aligned}
     \mathbf{h}_u^{(k)}&= Update^{(k)}(\mathbf{h}_u^{(k-1)}, \mathbf{m}_{\mathcal{N}(u)}^{(k)} + \sum_{v \in \mathcal{N}(u)} \mathbf{enc}_{v,\mathcal{N}(u)-v})\\
     &\hspace{-10pt}=  ReLU(\sum_{v\in \mathcal{N}(u)} \frac{1}{| \mathcal{N}(u) |} W^{(k)} \mathbf{h}^{(k)}_v + \sum_{v\in \mathcal{N}(u)}fc(Concat( \\
     &\hspace{15pt} \frac{1}{| \mathcal{N}(u) |} W^{(k)} \mathbf{h}^{(k)}_v,   (\mathbf{m}^{(k)}_{\mathcal{N}(u)} - \frac{1}{| \mathcal{N}(u) |} W^{(k)}, \mathbf{h}^{(k)}_v  ))) ,
\end{aligned}
\end{equation}
where $ReLU$ represenets the rectified linear activation function.

\begin{table*}[tbp]
\caption{Details of the six benchmark datasets used in our experiments.   } 

\centering  
\begin{tabular}{ l|c|cccccccc }
\toprule[1pt]
Task & Dataset  &Graphs &Nodes & Nodes/graph & Training & Validation & Test & Categories \\
\midrule[.6pt]

\multirow{2}{*}{Graph classification} &MNIST   &70K &4,939,668 &40-75  &55,000 &5000 &10,000 & 10 \\
                                       &CIFAR10 &60K &7,058,005 &85-150 &45,000 &5000 &10,000 & 10 \\

\midrule[.5pt]

\multirow{2}{*}{Node classification} &PATTERN & 14K &1,664,491 &44-188 &10,000 &2000 &2000 & 2 \\
&CLASTER & 12K &1,406,436 &41-190 &10,000 &1000 &1000 & 6 & \\

\midrule[.5pt]

Link prediction &TSP &12K &3,309,140 &50-500 &10,000 &1000 &1000 & 2  \\

\midrule[.5pt]

Graph regression &ZINC &12K &277,864 &9-37 &10,000 &1000 &1000 & n.a.  \\
\bottomrule[1pt]

\end{tabular}

\label{table:statics}
\end{table*}

\textbf{GatedGCN}.
A  GatedGCN layer takes a set of node features $\{\mathbf{h}_u^{(k-1)}, u\in V \}$ and a set of edge features $\{\mathbf{e}_{uv}^{(k-1)}, (u,v)\in E\}$ as input.
And it employs the edge gating mechanism \cite{marcheggiani2017encoding} to aggregate features passing from a node's local neighbours.
The message that passes from node $v \in \mathcal{N}(u)$ to node $u$ is defined as follows:
\begin{equation} \label{eq:loss_attn}
    \begin{aligned}
    \mathbf{m}^{(k)}_{\mathcal{N}(u)} &= \sum_{v\in \mathcal{N}(u)}\mathbf{m}_{v\to u}^{(k)}\\
                            &= \sum_{v\in \mathcal{N}(u)} {\alpha}_{uv}^{(k)} \odot ({F}^{(k)} {\mathbf{h}}_{v}^{(k)}),
    \end{aligned}
\end{equation}
where ${F}$ is trainable weight matrix, $\odot$ denotes the Hadamard product operator and $\mathbf{m}_{v\to u}^{(k)} = {\alpha}_{uv}^{(k)} \odot ({F}^{(k)} {\mathbf{h}}_{u}^{(k)})$ is the message that passes from node $v$ to node $u$, and $\alpha_{uv}^{(k)}$ is the gate vector and is defined as follows:
\begin{equation} \label{eq:loss_attn}
    \begin{aligned}
    &{\alpha}_{vu}^{(k)} = \frac{Sigmoid({\mathbf{e}'}_{vu}^{(k)})}{\sum_{v \in \mathcal{N}(u)}Sigmoid ( {\mathbf{e}'}^{(k)}_{vu})+ \epsilon} ,\\
    &{\mathbf{e}'}_{vu}^{(l)} = {A}\mathbf{h}_u^{(k-1)} +  {B}\mathbf{h}_u^{(k-1)} + {C}\mathbf{e}_{uv}^{(k-1)}
    \end{aligned}
\end{equation}
where ${A}^{(k)},{B}^{(k)},{C}^{(k)}$ are trainable weight matrices and $\epsilon$ is a small constant number.
With our neighbour-level message interaction encoding method, the embedding $\bm{h}_u^{(k)}$ is generated as follows:
\begin{equation} \label{eq:node_edge_feature_update}
    \begin{aligned}
    \mathbf{h}_u^{(k)} &= Update^{(k)}(\mathbf{h}_u^{(k-1)}, \mathbf{m}_{\mathcal{N}(u)}^{(k)} + \mathbf{enc}^{(k)}_{\mathcal{N}(u)})\\
     &= ReLU(BN({A}^{(k)} \mathbf{h}_{u}^{(k)}+ \mathbf{m}_{\mathcal{N}(u)}^{(k)} + \mathbf{enc}^{(k)}_{\mathcal{N}(u)})) + \mathbf{h}_{u}^{(k)}, \\
    \end{aligned}
\end{equation}
where  $BN$ denotes batch normalization, and  $\mathbf{enc}^{(k)}_{\mathcal{N}(u)}$ is defined as follows:
\begin{equation} \label{eq:aggr1}
    \begin{aligned}
     \mathbf{enc}_{\mathcal{N}(u)} &=  fc(Concat(\mathbf{m}^{(k)}_{v\to u}, \mathbf{m}_{\mathcal{N}(u)}^{(k)} - \mathbf{m}^{(k)}_{v\to u}))\\
     &=\sum_{v\in \mathcal{N}(u)}fc(Concat({\alpha}_{uv}^{(k)} \odot ({F}^{(k)} {\mathbf{h}}_{u}^{(k)}), \\ &\hspace{30pt} \mathbf{m}_{\mathcal{N}(u)} - \mathbf{\alpha}_{uv}^{(l)} \odot (\mathbf{F}^{(l)} {\mathbf{h}}_{u}^{(k)})))
    \end{aligned}
\end{equation}
The edge feature $\mathbf{e}_{uv}^{(k)}$ is updated as:

\begin{equation}
    \begin{aligned}
      \mathbf{e}_{uv}^{(k+1)} &= ReLU({\mathbf{e}'}_{ji}^{(k)}) + \mathbf{e}_{ji}^{(k)}
    \end{aligned}
\end{equation}

\fi

\iftrue
\section{Experiments}

We  evaluate our neighbour-level message interaction encoding method across vaious datasets,  demonstrating its effectiveness in  improving the model performance for  graph representation learning.

\begin{table*}[tbp]
\caption{Results of our graph convolutional network on MNIST and CIFAR10 on the graph superpixel classification task. The best results of our model and the baseline models are highlighted.}

\centering  
\begin{minipage}[b]{.4\linewidth}
\centering  
\begin{tabular}{l ccccc }
\toprule[1pt]
\multirow{2}{*}{ Model}   & \multicolumn{3}{c}{MNIST}  \\ 
\cmidrule(lr){2-4}
 &$K$ &Parameters &Accuracy ($\uparrow$) \\
\midrule[.6pt]
MLP & 4 & 104K &95.340$\pm$0.138 \\

GraphSAGE  \cite{hamilton2017inductive}   &4 &104K &97.312$\pm$0.097 \\
MoNet \cite{monti2017geometric} &4 &104K &90.805$\pm$0.032  \\

GAT       \cite{velickovic2018graph}  &4 &110K &95.535$\pm$0.205 \\

GIN  \cite{xu2019powerful}  &4  &105K &96.485$\pm$0.252 \\

RingGNN \cite{chen2019equivalence} &2  &105K &11.350$\pm$0.000  \\
                                   &2  &505K &91.860$\pm$0.449  \\
                                   &8  &506K &Diverged \\

3WLGNN \cite{maron2019provably} &3  &108K &95.075$\pm$0.961    \\
                                &3  &502K &95.002$\pm$0.419  \\
                                &8  &501K &Diverged           \\
\midrule[.8pt]
GCN \cite{kipf2017semi} &4  & 101K &90.705$\pm$0.218 \\

Ours + GCN \cite{kipf2017semi} &4 &357K &\textbf{91.725$\pm$0.270} \\

\midrule[.8pt]
GatedGCN   \cite{bresson2017residual}   &4 &104K &{\color{violet}97.340$\pm$0.143} \\
\hline
\multirow{3}{*}{\textbf{Ours + GatedGCN}} &4 &144K  &98.228$\pm$0.038   \\
                       &8  &284K  &98.512$\pm$0.048    \\
                       &12 &424K  &\textbf{98.635$\pm$0.049}  \\
\bottomrule[1.pt]

\end{tabular}
\end{minipage}
\begin{minipage}[b]{.4\linewidth}
\centering  
\begin{tabular}{lccccc }
\toprule[1pt]

\multirow{2}{*}{ Model}   & \multicolumn{3}{c}{CIFAR10}  \\ 
\cmidrule(lr){2-4}
 &$K$ &Parameters &Accuracy ($\uparrow$) \\
\midrule[.6pt]
MLP & 4 &104K &56.340$\pm$0.181 \\

GraphSAGE  \cite{hamilton2017inductive} &4&105K &65.767$\pm$0.308 \\

MoNet \cite{monti2017geometric}  &4&104K &54.655$\pm$0.518\\

GAT       \cite{velickovic2018graph}   &4&111K &64.223$\pm$0.455 \\

GIN  \cite{xu2019powerful}  &4&106K &55.255$\pm$1.527\\
RingGNN \cite{chen2019equivalence}  &2&105K &19.300$\pm$16.108\\
                                    &2&505K &39.165$\pm$17.114\\
                                    &8&510K &Diverged\\

3WLGNN \cite{maron2019provably}  &3&109K &59.175$\pm$1.593 \\
                                 &3&503K &58.043$\pm$2.512\\
                                 &8&502K &Diverged\\
\midrule[.8pt]
GCN \cite{kipf2017semi} &4&102K &55.710$\pm$0.381\\
Ours + GCN \cite{kipf2017semi} &4 &359K &\textbf{58.455$\pm$0.002}\\

\midrule[.8pt]
GatedGCN   \cite{bresson2017residual}    &4 &104K &{\color{violet}67.312$\pm$0.311}  \\
\hline
\multirow{3}{*}{\textbf{Ours + GatedGCN}} &4 &144K  &75.001$\pm$0.257\\
                        &8 &284K  &76.395$\pm$0.101  \\
                        &12 &424K &\textbf{76.515$\pm$0.081}  \\
\bottomrule[1.pt]
\end{tabular}
\end{minipage}

\label{table:results_graphclassification}
\end{table*}

\subsection{Datasets and Setup}
\textbf{Datasets}.
The proposed neighbour-level message encoding method is experimentally validated on four domain tasks on the following six  benchmark datasets \cite{dwivedi2020benchmarking}. 
\begin{itemize}
  \item \textbf{MNIST} and \textbf{CIFAR-10}.
      The two dataets are  superpixel graph datasets, and we use the two datasets for the graph classification task.
      The superpixel graphs are extracted from the  images of the  MNIST  dataset \cite{lecun1998gradient} and CIFAR10 dataset \cite{krizhevsky2009learning} using the SLIC method \cite{achanta2012slic}.
      The superpixels represent small regions of homogeneous intensity in images.
      Samples of the superpixel graphs in MNIST and CIFAR10 are shown in Figure \ref{fig:samples-mnist-cifar}.

  \item \textbf{PATTERN} and \textbf{CLUSTER}.
     The two datasets contain 14K and 12K graphs respectively and are used for the node classification task.
     The PATTERN dataset is used for validating the model's performance for recognizing subgraphs, and the CLUSTER dataset is used for validating the model's performance for recognizing community graphs.

  \item \textbf{TSP}. The  TSP dataset is constructed to evaluate graph convolutional networks on solving the travelling salesman problem (TSP), which is a classical NP-Hard combinatorial problem.
      We validate if or not the predicted edges of our  model belong to the optimal TSP solution obtained using the Concorde solver \cite{applegate2006concorde}.

   \item \textbf{ZINC}. This dataset which contains 12K samples is a subset of the ZINC molecular graph (250K) dataset \cite{irwin2012zinc}.
    We evaluate the performance of our model for regressing the molecular property (or called constrained solubility) on this dataset.
\end{itemize}

The details of the six benchmark datasets are reported in Table \ref{table:statics}.

\textbf{Evaluation metrics}.
Following the work of Dwivedi et al. \cite{dwivedi2020benchmarking}, the following metrics are utlized for performance evaluation according to different tasks.
\begin{itemize}
  \item \textbf{Accuracy}. For the superpixel graph classification task on MNIST and CIFAR10, classification accuracy is used for evaluating the model performance.
      For the node classification task on PATTERN and CLUSTER, weighted accuracy is used for evaluating the model performance.

  \item  \textbf{F1 score}.
       For the edge prediction task on TSP, due to high class imbalance, i.e., only the edges in the TSP tour are labeled as positive, the model performance is reported using the F1 score for the positive class. 

  \item \textbf{MAE}.  The MAE (mean absolute error) is used for evaluating the model's  performance for regressing the molecular property on the ZINC dataset.
\end{itemize}

\textbf{Baseline models}. 
To demonstrate the superior performance of our graph  network model, we compare our results against a diverse of baseline models including MLP (multilayer perceptron), GCN \cite{kipf2017semi}, GraphSAGE  \cite{hamilton2017inductive}, MoNet \cite{monti2017geometric}, GAT       \cite{velickovic2018graph}, GatedGCN   \cite{bresson2017residual}, GIN  \cite{xu2019powerful}, RingGNN \cite{chen2019equivalence} and 3WLGNN \cite{maron2019provably}.

\begin{table*}[tbp]
\caption{Results of our graph convolutional network  on the PATTERN dataset  and CLUSTER dataset on node classification. The best result of the baseline models is highlighted in violet.}

\centering  
\begin{minipage}[b]{.4\linewidth}
\centering  
\begin{tabular}{l ccccc }
\toprule[1pt]
\multirow{2}{*}{ Model}   & \multicolumn{3}{c}{PATTERN}  \\ 
\cmidrule(lr){2-4}
 &$K$ &Parameters &Accuracy ($\uparrow$) \\
\midrule[.6pt]
MLP &4  &105K &50.519$\pm$0.000   \\


GraphSAGE  \cite{hamilton2017inductive}    &4 &102K &50.516$\pm$0.001  \\
                                           &16 &503K &50.492$\pm$0.001  \\

MoNet \cite{monti2017geometric} &4 &104K &85.482$\pm$0.037   \\
                                &16 &511K &85.582$\pm$0.038   \\
GAT       \cite{velickovic2018graph}  &4 &110K &75.824$\pm$1.823  \\
                                      &16 &527K &78.271$\pm$0.186  \\

GIN  \cite{xu2019powerful}  &4 &101K &85.590$\pm$0.011   \\
                             &16 &509K &85.387$\pm$0.136  \\
RingGNN \cite{chen2019equivalence} &2 &105K &{\color{violet}86.245$\pm$0.013} \\
                                    &2 &505K &86.244$\pm$0.025  \\
                                    &8 &506K &Diverged  \\

3WLGNN \cite{maron2019provably} &3 &104K &85.661$\pm$0.353  \\
                                &3 &503K &85.341$\pm$0.207  \\
                                &8 &582K &Diverged  \\
\midrule[.8pt]
GCN \cite{kipf2017semi} &4  &101K &63.880$\pm$0.074  \\
Ours + GCN \cite{kipf2017semi} &4 &358K &\textbf{85.615$\pm$0.049} \\

\midrule[.8pt]
GatedGCN   \cite{bresson2017residual}   &4 &104K &84.480$\pm$0.122  \\
                                        &16 &502K &85.568$\pm$0.088 \\
\hline
\multirow{4}{*}{\textbf{Ours + GatedGCN}}
                       & 4 &143K  &85.669$\pm$0.034     \\
                       &8  &283K  &86.336$\pm$0.136    \\
                       &12 &423K  &86.601$\pm$0.084    \\
                       &16  &563K  &\textbf{86.643$\pm$0.047}     \\

\bottomrule[1.pt]

\end{tabular}
\end{minipage}
\begin{minipage}[b]{.4\linewidth}
\centering  
\begin{tabular}{lccccc }
\toprule[1pt]

\multirow{2}{*}{ Model}   & \multicolumn{3}{c}{CLUSTER}  \\ 
\cmidrule(lr){2-4}
 &$K$ &Parameters &Accuracy ($\uparrow$) \\
\midrule[.6pt]
MLP    &4 &106K &20.973$\pm$0.004\\

GraphSAGE  \cite{hamilton2017inductive}    &4 &102K &50.454$\pm$0.145 \\
                                            &16 &503K &63.844$\pm$0.110 \\

MoNet \cite{monti2017geometric}  &4  &104K &58.064$\pm$0.131\\
                                 &16 &512K &66.407$\pm$0.540\\
GAT       \cite{velickovic2018graph}   &4 &111K &57.732$\pm$0.323\\
                                      &16  &528K &70.587$\pm$0.447\\

GIN  \cite{xu2019powerful}   &4 &104K &58.384$\pm$0.236\\
                              &16 &518K &64.716$\pm$1.553\\
RingGNN \cite{chen2019equivalence}  &2 &105K &42.418$\pm$20.063\\
                                     &2 &524K &22.340$\pm$0.000\\
                                     &8 &514K  &Diverged\\

3WLGNN \cite{maron2019provably}  &3 &106K &57.130$\pm$6.539\\
                                 &3 &507K &55.489$\pm$7.863\\
                                 &8 &587K  &Diverged\\
\midrule[.8pt]
GCN \cite{kipf2017semi}  &4 &102K &53.445$\pm$2.029\\
Ours + GCN \cite{kipf2017semi} &4 &359K &\textbf{58.313$\pm$0.196} \\
\midrule[.8pt]

GatedGCN   \cite{bresson2017residual}   &4 &104K &60.404$\pm$0.419\\
                                        &16 &503K &{\color{violet}73.840$\pm$0.326}\\
\hline

\multirow{4}{*}{\textbf{Ours + GatedGCN}}  &4 &144K  &63.212$\pm$0.130   \\
                                 &8 &284K  &73.055$\pm$0.135  \\
                                 &12 &424K &75.458$\pm$0.108  \\
                                 &16 &564K &\textbf{76.163$\pm$0.101} \\
\bottomrule[1.pt]
\end{tabular}
\end{minipage}

\label{table:results_nodeclassification}
\end{table*}

\textbf{Implementation details}.
We closely follow the implementation details as the work of Dwivedi \cite{dwivedi2020benchmarking}.
We adopt Adam algorithm \cite{kingma2014adam} for optimizing our graph network  model.
We set the initial learning rate to 0.0001 and divide the learning rate by 2 when the validation loss has not been decreased for 10 or 20 or 30 epochs.
When the learning rate is decreased to a value less than $10^{-6}$, we terminate the training procedure.
We also apply the SSFG regularization method \cite{zhang2022ssfg} in our model.
Our model is implemented in  Pytorch \cite{paszke2017automatic} with the DGL library \cite{wang2019deep}.
The experiments are carried out on a 24GB memory GPU.
We validate our  model using various graph convolutional layers, {e.g.,} 4, 8, 12, 16.
For each experiment, we validate our model for 4 times using different random seeds and report the mean and standard deviation over the 4 runs.

\begin{table}[tbp]
\caption{Experimental results on the TSP dataset on link prediction. OOM denotes out of memory.}

\centering  

\begin{tabular}{lccccc }
\toprule[1pt]

\multirow{2}{*}{ Model} &   & \multicolumn{2}{c}{TSP}  \\
\cmidrule(lr){2-4}
 &$K$ &Parameters &  F1 ($\downarrow$)  \\
\midrule[.8pt]
MLP & 4 &97K & 0.544$\pm$0.0001  \\

GraphSAGE  \cite{hamilton2017inductive}        &4      & 90K &0.665$\pm$0.003  \\

MoNet \cite{monti2017geometric}    &4 &99K &0.641$\pm$0.002 \\

GAT       \cite{velickovic2018graph}  &4 & 96K &0.671$\pm$0.002\\

GIN  \cite{xu2019powerful}  &4 &99K &0.656$\pm$0.003\\

RingGNN \cite{chen2019equivalence} &2 &107K &0.643$\pm$0.024\\
                                   &2 &508K &0.704$\pm$0.003\\
                                   &8 &506K &Diverged\\
3WLGNN \cite{maron2019provably} &3 &106K &0.694$\pm$0.073 \\
                                &3 &507K &0.288$\pm$0.311 \\
                                &8 &509K &OOM \\
\midrule[.8pt]
GCN \cite{kipf2017semi} &4 &96K &0.630$\pm$0.0001\\

Ours + GCN &4 &269K &\textbf{0.651$\pm$0.0002} \\

\midrule[.8pt]
GatedGCN   \cite{bresson2017residual}   &4 &98K &{\color{violet}0.791$\pm$0.003}\\
\hline

\multirow{4}{*}{\textbf{Ours + GatedGCN}} &4 &132K  &0.811$\pm$0.001 \\
              &8  &253K  &0.832$\pm$0.001 \\
              & 12  &373K  &0.841$\pm$0.001 \\
              & 16  &495K  &\textbf{0.844$\pm$0.001} \\

\bottomrule[1pt]

\end{tabular}

\label{table:tsp}
\end{table}

\begin{table}[tbp]
\caption{Experimental results  on the ZINC dataset on the graph regression task. Note that  only node features are used for training our model.}

\centering  

\begin{tabular}{lccccc }
\toprule[1pt]

\multirow{2}{*}{ Model} &\multirow{2}{*}{$K$}   & \multicolumn{2}{c}{ZINC}  \\
\cmidrule(lr){3-4}
 & &Parameters &  MAE ($\downarrow$)  \\
\midrule[.8pt]
MLP & 4 &109K & 0.706$\pm$0.006  \\

GraphSAGE  \cite{hamilton2017inductive}        &4      & 95K &0.468$\pm$0.003  \\
                                               &16    &505K &0.398$\pm$0.002\\
MoNet \cite{monti2017geometric}    &4 &106K &0.397$\pm$0.010 \\
                                   &16 &504K &{\color{violet}0.292$\pm$0.006} \\
GAT       \cite{velickovic2018graph}  &4 & 102K &0.475$\pm$0.007\\
                                      &16 &531K &0.384$\pm$0.007\\

GIN  \cite{xu2019powerful}  &4 &103K &0.387$\pm$0.015\\
                            &16 &510K &0.526$\pm$0.051\\
RingGNN \cite{chen2019equivalence} &2 &98K &0.512$\pm$0.023\\
3WLGNN \cite{maron2019provably} &3 &102K &0.407$\pm$0.028 \\

\midrule[.8pt]
GCN \cite{kipf2017semi} &4 &103K &0.459$\pm$0.006\\

Ours + GCN  &4 &356K &\textbf{0.375$\pm$0.002}\\

\midrule[.8pt]
GatedGCN   \cite{bresson2017residual}   &4 &106K &0.435$\pm$0.011\\
GatedGCN-E \cite{bresson2017residual} & 4 &106K & 0.375$\pm$0.003 \\
\hline
\multirow{3}{*}{\textbf{Ours + GatedGCN}} &4 &146K  &0.268$\pm$0.005 \\
              &8  &285K  &0.233$\pm$0.004 \\
              & 12  &425K  &\textbf{0.226$\pm$0.002} \\

\bottomrule[1pt]

\end{tabular}

\label{table:zinc}
\end{table}

\subsection{Experimental Results}
\ifshowfig
\begin{figure*}[!t]

  \vspace{5pt}
  \begin{center}
  \includegraphics[width=0.97\textwidth]{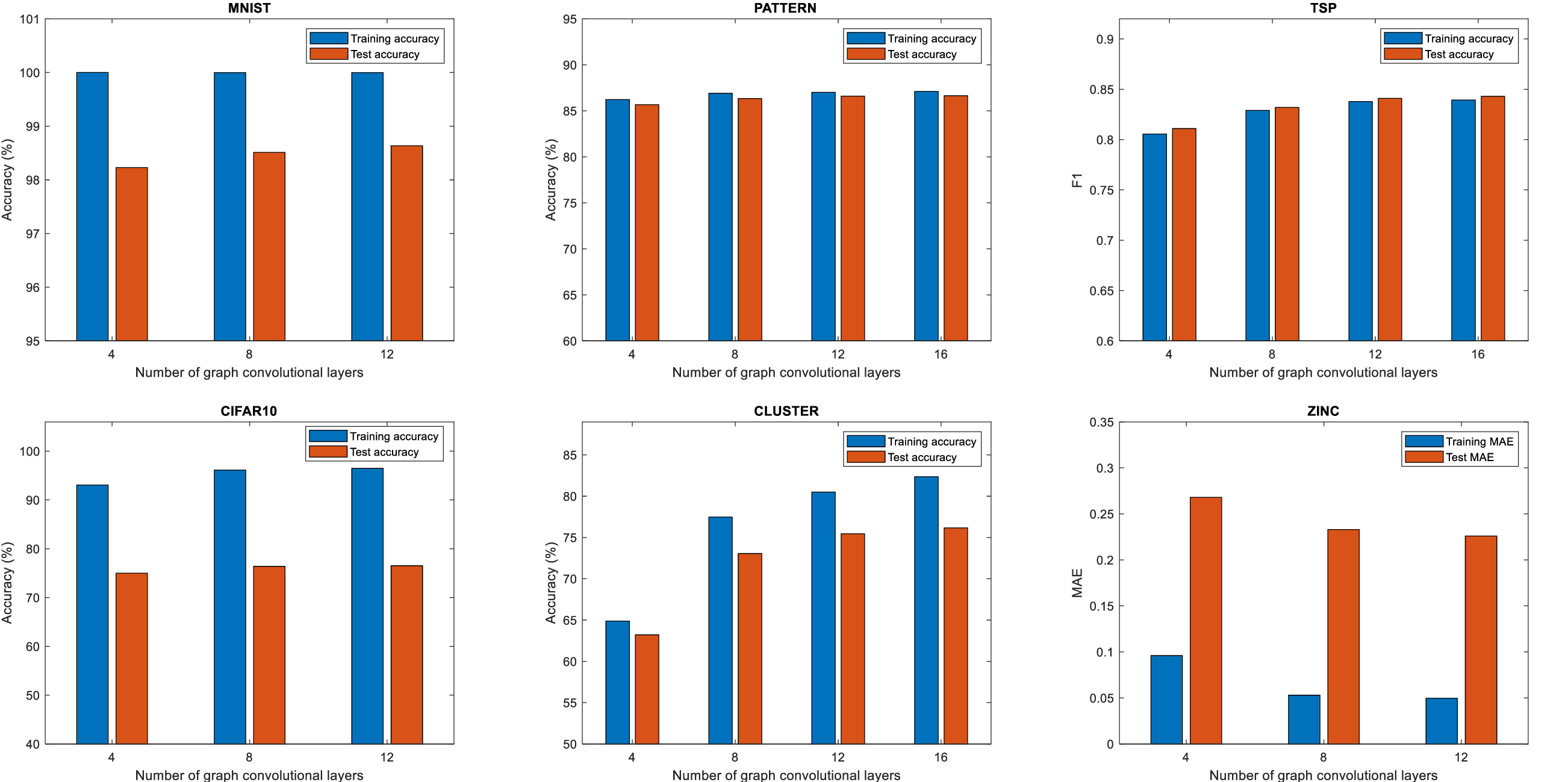} 
  \end{center}
  \caption{Performance our model using different graph convolutional layers.}
  \label{fig:accuracy}
\end{figure*}
\fi

\begin{table*}[tbp]
\caption{Ablation study: importance of  each feature component for generating a node's embedding (see Eq. (\ref{eq:node_edge_feature_update})) on the model performance. }
\centering
\begin{tabular}{l|cc cc |c|c }
\toprule[1pt]

\multirow{2}{*}{Method} & \multicolumn{4}{c|}{Accuracy ($\uparrow$)} &\multicolumn{1}{c|}{F1 ($\uparrow$) } & \multicolumn{1}{c}{MAE ($\downarrow$) } \\
  &MNIST ($K$=12) &CIFAR10 ($K$=12) &PATTERN ($K$=16) &CLUSTER ($K$=16) &TSP ($K$=16) &ZINK ($K$=12)  \\

\midrule[.8pt]
${A} \mathbf{h}_{u}+\mathbf{m}_{\mathcal{N}(u)}$ &98.140  &72.032 &85.723 &75.245 &0.818 &0.281\\  
${A} \mathbf{h}_{u}+\mathbf{enc}^{(k)}_{\mathcal{N}(u)}$ &98.490  &76.383 &86.589 &76.158 &0.840 &0.227\\
$\mathbf{m}_{\mathcal{N}(u)}+\mathbf{enc}_{\mathcal{N}(u)}$ & 98.573 &76.497 &86.620 &76.056 &0.843 &0.226\\
\midrule[.5pt]
${A} \mathbf{h}_{u} + \mathbf{m}_{\mathcal{N}(u)}+\mathbf{enc}_{\mathcal{N}(u)}$ &98.635 &76.515 &86.643 &76.163 &0.844 &0.226\\

\bottomrule[1pt]

\end{tabular}
\label{table:ablation_study}
\end{table*}

\textbf{Graph classification}.
For the superpixel graph classification task on MNIST and CIFAR10, we conduct experiments using three different graph convolutional layers, {i.e.}, $K\in\{4,8,12\}$.
The quantitative results are reported in Table \ref{table:results_graphclassification}, comparing our graph network model against the baseline models.
We see that our model which uses GatedGCN as the base network and with 4 graph convolutional layers obtains 98.228\% and 75.001\% accuracy on MNIST and CIFAR10, respectively.
Further increasing the number of graph convolutional layers to 8 and 12 results in further performance gains.
The vanilla GatedGCN achieves improved performance on the two datasets compared to the other baseline models.
When compared to the vanilla GatedGCN,  our 12 graph convolutional layer model  improves the model performance by 1.295\%   and 9.203\% on the two datasets respectively.
As far as we know, our graph network nodel achieves better  performance than previous models on the two datasets.

\textbf{Node classification}.
For  experiments  on the PATTERN dataset and CLUSTER dataset, our graph network model is validated using  four different graph convolutional layers, i.e., $K=4,8,12,16$.
Table \ref{table:results_nodeclassification} compares the results of our model against the baseline models.
For the GatedGCN as the base network, our 16 graph convolutional layer model  achieves 86.643\% and 76.163\% weighted accuracy on PATTERN and CLUSTER, respectively.
It can be seen that our model outperforms all the baseline models.
RingGNN \cite{chen2019equivalence} achieves the best performance among the baseline models on PATTERN.
When compared to RingGNN, our model achieves 0.398\% improved performance  on PATTERN.
The vanilla GatedGCN   \cite{bresson2017residual} obtains improved performance compared to the other baseline models on the CLUSTER dataset.
Compared with the vanilla GatedGCN, applying our encoding methods yields a 2.323\% performance improvement.

\textbf{Link prediction}.
The experimental results on link prediction on the TSP dataset is reported in Table \ref{table:tsp}.
We see that our graph model which uses the GateGCN as the base network outperforms all the baseline models by a large margin, demonstrating the effectiveness of the proposed neighbour-level message interaction encoding method for improving graph convolutional network models  for link prediction.
The vanilla GatedGCN achieves an F1 score of 0.791, which is the best result among the baseline models.
Our model with $K=16$ outperforms the vanilla GatedGCN by 0.053.
Once again, applying out method advances the state of the art performance.

\textbf{Graph regression}.
The results  on the ZINC dataset on the graph regression task are reported Table \ref{table:zinc}.
Note that during the training of our model, only  node features (i.e., types of heavy atoms) are utilized. 
Compared with the vanilla GatedGCN-E \cite{bresson2017residual} , which uses both node features and edge features (i.e., bond types), our 12 layer graph network model which uses GatedGCN as the base network  results in a 0.149 reduced MAE. 
The MoNet model \cite{monti2017geometric} yields  the best result among the baseline models.
Compared with MoNet \cite{monti2017geometric},  our 12 layer graph network model which uses GatedGCN as the base network reduces the MAE from 0.292 to 0.226.
The results show that integrating neighbour-level message interaction information also  improves the representation learning performance for graph regression.

Figure \ref{fig:accuracy} shows the training and test accuracies/MAEs/F1 scores of our model which uses GatedGCN as the base network with different graph convolutional layers on the benchmark datasets.
It can be seen that the training and test performances are improved consistently  as the number of graph convolutional layers increases.

\textbf{Ablation study}.
In a layer of our model which uses GatedGCN as the base network, three features, i.e., ${A}  \mathbf{h}_{u} $, $\mathbf{m}_{\mathcal{N}(u)}$ and $\mathbf{enc}_{\mathcal{N}(u)}$ are used to update node embeddings (see Equation (\ref{eq:node_edge_feature_update})).
We carry out an ablation study to show the significance of each feature on the overall model performance.
Note when only ${A} \mathbf{h}_{u}$ and $\mathbf{m}_{\mathcal{N}(u)}$ are used, our graph  network model is equivalent to the vanilla GatedGCN model.
As shown in Table \ref{table:ablation_study}, we see that integrating the neighbour-level message interaction information consistently improves the graph representation learning performance on all the benchmark datasets.
It can also be seen that the use of $\mathbf{enc}_{\mathcal{N}(u)}$ performs better than the use of $\mathbf{m}_{\mathcal{N}(u)}$.
The results demonstrate that encoding  encoding neighbour-level message interaction information is a generic method that consistently improves the representation learning performance on graphs.

\fi

\iftrue
\section{Conclusion}
This paper proposed a neighbour-level message interaction information encoding method for improving graph convolutional  network.
At each layer of message passing, we first follow the conventional framework to generate an aggregated message for a node through aggregating information from the node's local neighbourhood.
For each of the node's local neighbours, we learn an encoding between the message from the neighbour node and the aggregated message from the rest neighbour nodes using a fully connected layer.
These learned encodings are aggregated to generate a neighbour-level message interaction encoding.
Finally, the sum of the aggregated message and neighbour-level message encoding is taken to generate an updated embedding for the  node.
Through this way, neighbour-level message interaction information is  integrated into the generated node embeddings, and therefore the representational ability of the generated embeddings is improved.
We experimentally evaluated our graph network model on four graph-based tasks, including superpixel graph classification, node classification, edge prediction and graph regression, on six recently released benchmark datasets, including MNIST, CIRAR10, PATTER, CLUSTER, TSP and ZINC.
The results showed that integrating the proposed neighbour-level message interaction  information  encoding method is a generic approach  to improve the representation learning performance on graph-structured data.
We demonstrated that integrating neighbour-level message interaction information achieves new  state-of-the-art performance on the benchmark datasets.

\section{Acknowledgement}
The authors would like to sincerely thank the editors and reviewers for  on reviewing this manuscript.
\fi

\bibliographystyle{IEEEtran}
\bibliography{mybib}

\end{document}